# tmVar 3.0: an improved variant concept recognition and normalization tool

Chih-Hsuan Wei[1], Alexis Allot[1], Kevin Riehle[2], Aleksandar Milosavljevic[2] and Zhiyong Lu[1*]

[1]National Center for Biotechnology Information (NCBI), National Library of Medicine (NLM), National Institutes of Health (NIH), Bethesda, MD 20894, USA

[2]Department of Molecular and Human Genetics, Baylor College of Medicine, Houston, Texas, United States of America

**Abstract**

Previous studies have shown that automated text-mining tools are becoming increasingly important for successfully unlocking variant information in scientific literature at large scale. Despite multiple attempts in the past, existing tools are still of limited recognition scope and precision. We propose tmVar 3.0: an improved variant recognition and normalization tool. Compared to its predecessors, tmVar 3.0 is able to recognize a wide spectrum of variant related entities (e.g., allele and copy number variants), and to group different variant mentions belonging to the same concept in an article for improved accuracy. Moreover, tmVar3 provides additional variant normalization options such as allele-specific identifiers from the ClinGen Allele Registry. tmVar3 exhibits a state-of-the-art performance with over 90% accuracy in F-measure in variant recognition and normalization, when evaluated on three independent benchmarking datasets. tmVar3 is freely available for download. We have also processed the entire PubMed and PMC with tmVar3 and released its annotations on our FTP.

**Availability:** ftp://ftp.ncbi.nlm.nih.gov/pub/lu/tmVar3
**Contact:** zhiyong.lu@nih.gov

## Introduction

Genomic variant is one of the concepts of key relevance for precision medicine which aims to support personalized treatments based on an individual's genetic profile. To better understand the mechanism of genetic diseases, (semi-)automatically collecting and assimilating published knowledge about sequence variants in scientific literature becomes an increasingly important task. A recent study (Lee, et al., 2021) reviewed a number of existing software tools previously developed for such a task (Caporaso, et al., 2007; Cejuela, et al., 2017; Cheng, et al., 2020; Thomas, et al., 2016; Wei, et al., 2017). Most of the tools are developed by regular expression based on the HGVS nomenclature and frequent variant forms in text. We previously developed tmVar (Wei et al., 2013), which uses a machine-learning based approach to optimally recognize variant components (wild type, mutant, and position). More recently, a new function was added to tmVar so it performs variant normalization by linking recognized variant mentions to standard concept identifiers (Wei, et al., 2017). Specifically, tmVar 2.0 normalizes variant mentions to dbSNP (https://www.ncbi.nlm.nih.gov/snp/) RS identifiers like many other tools (e.g., Thomas et al., 2016). Relying on the variant linking results of tmVar, several downstream text mining applications were successfully developed (Allot, et al., 2018; Nie, et al., 2019).

Despite these efforts, existing variant extraction tools are still limited in (1) recognizing variant types of a broader scope such as incomplete variants (e.g., V600) and related concepts (e.g., genomic regions). Such concepts are found to play important roles in connecting variants with disease and drug information in the same article. (2) linking mentions to specific alleles. Note that dbSNP RS identifiers only record the polymorphism at a specific position (e.g., rs113488022) but do not differentiate specific (e.g., T>A vs. T>C) registered in the ClinGen Allele Registry (CAR) (Pawliczek, et al., 2018). We herein propose a new comprehensive variant extraction system that specifically addresses all these challenges by improving our previous tmVar tool. The improved tmVar 3.0 system achieves consistent high precision and recall on several publicly available gold standard corpora and is freely available for the scientific community.

**Methods**

We first expanded the recognition scope to cover multiple variant related concepts which were rarely discussed in previous studies, i.e., incomplete variant mentions (e.g., Cys326; guanine to cytosine), copy number variants (e.g., chr19:54,666,173-54,677,766 bp del), reference sequence (RefSeq) (e.g., NM_203475.1), chromosomal locations (e.g., chromosome 5 q 33) and genomic regions (e.g., chr7:156583796-156584569) as shown in Table 1.

**Table 1.** The mutation types extracted by tmVar3 and the examples.

| Type | Example |
| --- | --- |
| SNP | Rs763780 |
| DNA Mutation | c.1976A>T |
| DNA Allele | 1976A |
| DNA Change | A>T |
| Protein Mutation | p.Gln659Leu |
| Protein Allele | glutamine at codon 659 |
| Protein Change | methionine to threonine |
| Other Mutations | 306 base pair insertion |
| Copy number variant | Chr 15: 3,18,33,000-3,74,77,000bp deletion |
| RefSeq | NM_203475.1 |
| Chromosome | 10q11.12 |
| Genomic Region | Chr10: 46123781-51028772 |

To better support variant-related text mining research (e.g., mining variant-disease associations), tmVar 3.0 groups the variants which are the same but in different form/types (e.g., DNA and protein variants). For instance, in PMID: 20577006, we group the variants (i.e., P799L and P799) belonging to rs121912637. In this article, P799 co-occurs with disease *metatropic dysplasia* in the same sentence, but not P799L. In this case, grouping the two variant mentions makes it easier to link P799L to the correct disease.

Third, tmVar3 provides alterative options to normalize a particular variant. In addition to providing RS ids that record all the possible allele changes on a specific genomic position, we offer three allele-specific options for improved precision in the normalization results:

(1) ClinGen Allele Registry (CAR) Canonical Allele Identifier (CA ID) (e.g., CA16602736) and (2) the combination of a RS ID and the specific allele (e.g., rs113488022(T>A)). Furthermore, we observed that more than half of the variant mentions cannot be linked to an existing record in dbSNP or CAR databases. In such cases, tmVar 3.0 finds the corresponding gene of the variant in the text and normalizes it with the variant as the third option (e.g., BRAF: c.1799T>A).

Finally, in tmVar3, we improved our recognition algorithm on some previously difficult edge cases such as variants described in natural language (e.g., "nine-nucleotide deletion starting at position 1952"), or with a missing space between the gene and variant (e.g., "BRAFV600E").

**Results**

The newly improved tmVar3 system is assessed on three separate benchmarking datasets (i.e., OSIRIS (Bonis, et al., 2006), Thomas (Thomas, et al., 2016) and out tmVar corpus (Wei, et al., 2013)). The evaluation results of tmVar3 are shown in Table 2 and compared with SETH (Thomas, et al., 2016), a previous state-of-the-art method producing normalized dbSNP RS IDs. As can be seen, tmVar3 achieves consistently higher accuracy (over 90% in F-measure) than SETH on the three public corpora. To facilitate the use of tmVar results at PubMed scale, we have processed the entire PubMed/PMC open access and incorporated the results in the NCBI web server PubTator (Wei, et al., 2016). The annotations are also freely available via FTP.

**Table 2.** tmVar 3.0 performance comparison with SETH on three public benchmarking datasets: tmVar (Wei, et al., 2013), ORIRIS (Bonis, et al., 2006), and Thomas (Thomas, et al., 2016) for variant normalization.

| Corpus | Method | Precision | Recall | F-score |
|---|---|---|---|---|
| tmVar | tmVar3 | 96.99% | 91.71% | 94.28% |
| | SETH | 86.51% | 69.91% | 77.33% |
| OSIRIS | tmVar3 | 97.72% | 84.58% | 90.68% |
| | SETH | 94.21% | 69.38% | 79.91% |
| Thomas | tmVar3 | 91.01% | 90.32% | 90.67% |
| | SETH | 95.58% | 57.50% | 71.80% |

**Conclusion**

We introduced tmVar 3.0, an improved open-source software tool with a broader scope and better accuracy for variant concept recognition and normalization, compared to its predecessors. tmVar 3.0 can recognize most of the variants even when the variants are described with partial information (e.g., amino acid change without the sequence position) or with natural language. tmVar 3.0 groups different mentions of the same variant together based on the context for improved normalization performance. As a result, tmVar3 achieves superior variant recognition and normalization. In the future, we would like to further enhance and expand tmVar by better linking variants with other closely related concepts such as drugs and diseases.

**Funding**

This work was supported by the National Institutes of Health intramural research program, National Library of Medicine and in part by the NIH NHGRI Clinical Genome Resource (ClinGen) grant U24 HG009649.

**References**

Allot, A., et al. (2018) LitVar: a semantic search engine for linking genomic variant data in PubMed and PMC, Nucleic Acids Res., 46, W530–W536.

Bonis, J., Furlong, L.I. and Sanz, F. (2006) OSIRIS: a tool for retrieving literature about sequence variants, Bioinformatics, 22, 2567-2569.

Caporaso, J.G., et al. (2007) MutationFinder: a high-performance system for extracting point mutation mentions from text, Bioinformatics, 23, 1862-1865.

Cejuela, J.M., et al. (2017) nala: text mining natural language mutation mentions, Bioinformatics, 33, 1852-1858.

Cheng, C., Tan, F. and Wei, Z. (2020) DeepVar: An end-to-end deep learning approach for genomic variant recognition in biomedical literature. Proceedings of the AAAI Conference on Artificial Intelligence. pp. 598-605.

Lee, K., Wei, C.-H. and Lu, Z. (2021) Recent advances of automated methods for searching and extracting genomic variant information from biomedical literature, Briefings in bioinformatics, 22, bbaa142.

Nie, A., et al. (2019) LitGen: Genetic literature recommendation guided by human explanations. PACIFIC SYMPOSIUM ON BIOCOMPUTING 2020. World Scientific, pp. 67-78.

Pawliczek, P., et al. (2018) ClinGen Allele Registry links information about genetic variants, Human mutation, 39, 1690-1701.

Thomas, P., et al. (2016) SETH detects and normalizes genetic variants in text, Bioinformatics, 32, 2883-2885.

Wei, C.-H., et al. (2013) tmVar: A text mining approach for extracting sequence variants in biomedical literature, Bioinformatics, 29, 1433-1439.

Wei, C.-H., Leaman, R. and Lu, Z. (2016) Beyond accuracy: creating interoperable and scalable text-mining web services, Bioinformatics, 32, 1907-1910.

Wei, C.-H., et al. (2017) tmVar 2.0: integrating genomic variant information from literature with dbSNP and ClinVar for precision medicine, Bioinformatics, 34, 80-87.